\def\year{2020}\relax
\documentclass[letterpaper]{article} 
\usepackage{aaai20}  
\usepackage{times}  
\usepackage{helvet} 
\usepackage{courier}  
\usepackage[hyphens]{url}  
\usepackage{graphicx} 
\usepackage{graphicx}  
\usepackage[utf8]{inputenc} 
\usepackage[T1]{fontenc}    
\usepackage{url}            
\usepackage{booktabs}       
\usepackage{amsfonts}       
\usepackage{nicefrac}       
\usepackage{microtype}      
\usepackage{mathtools}
\usepackage{amsmath, amssymb}
\usepackage[capitalize]{cleveref}
\usepackage{subcaption}
\usepackage{algorithm}
\usepackage{algorithmic}
\usepackage{color}

\newcommand{\data}{x}
\newcommand{\Data}{X}
\newcommand{\labelmat}{\boldsymbol{Y}}
\newcommand{\labelvec}{\boldsymbol{y}}
\newcommand{\numdata}{n}
\newcommand{\numconstraints}{m}
\newcommand{\numclasses}{C}
\newcommand{\weakvec}{\boldsymbol{q}}

\newcommand{\params}{\theta}
\newcommand{\lagrange}{\gamma}
\newcommand{\lagrangevec}{\boldsymbol{\lagrange}}
\newcommand{\loss}{L}
\newcommand{\stepsize}{\rho}

\newcommand{\reals}{\mathbb{R}}

\title{Stochastic Generalized Adversarial Label Learning}
\author{
Chidubem Arachie\\
Department of Computer Science\\
Virginia Tech\\
\texttt{achid7@vt.edu} \\
\And
Bert Huang \\
Department of Computer Science\\
Virginia Tech \\
\texttt{bhuang@vt.edu} \\
}

\begin{document}

\maketitle

\begin{abstract}

The usage of machine learning models has grown substantially and is spreading into several application domains. A common need in using machine learning models is collecting the data required to train these models. In some cases, labeling a massive dataset can be a crippling bottleneck, so there is need to develop models that work when training labels for large amounts of data are not easily obtained. A possible solution is weak supervision, which uses noisy labels that are easily obtained from multiple sources. The challenge is how best to combine these noisy labels and train a model to perform well given a task.  
In this paper, we propose stochastic generalized adversarial label learning (Stoch-GALL), a framework for training machine learning models that perform well when noisy and possibly correlated labels are provided. Our framework allows users to provide different weak labels and multiple constraints on these labels. Our model then attempts to learn parameters for the data by solving a non-zero sum game optimization. The game is between an adversary that chooses labels for the data and a model that minimizes the error made by the adversarial labels.  We test our method on three datasets by training convolutional neural network models that learn to classify image objects with limited access to training labels. Our approach is able to learn even in settings where the weak supervision confounds state-of-the-art weakly supervised learning methods. The results of our experiments demonstrate the applicability of this approach to general classification tasks.
\end{abstract}

\section{Introduction}

Recent success of deep learning has seen an explosion in interest towards building large-scale models for various applications. Training deep models often involves using massive amounts of training data whose labels are not easily obtained or available.
Collecting labeled data for training these large-scale models is a major bottleneck since these labels are usually provided by expert annotators and can be expensive to gather. Weak supervision offers an alternative for training machine learning models because it relies on approximate labels that are easily obtained.  Weakly supervised learning alleviates some of the difficulties and cost associated with supervised learning by only requiring annotators to provide rules or approximate indicators that automatically label the data. Ideally, these annotators or human experts provide several weak rules in the form of feature annotations, heuristic patterns, or programmatically generated labels. The difficulty lies in combining multiple weak supervision signals that may be from various sources, make dependent errors, or sometimes have different noise levels within them. Researchers have developed methods \cite{ratner2017snorkel,ratner2016data,arachie2019adversarial} to combine the different weak supervision signals to train models robust to redundancies and errors in the weak supervision.  These approaches are limited in ways such as being overly optimistic about the independence of weak signals, being only able to handle binary classification, or having poor scalability. We address these limitations in this paper.

We develop a general framework that encodes multiple linear constraints on the weak supervision signals. Our algorithm learns from the weak supervision by solving a non-zero sum game between an adversary and the model. By using a non-zero sum game, our formulation provides more flexibility than previous methods and allows for different loss functions to be used. Thus, we enable learning for other forms such as multiclass classification, multilabel classification, and structured prediction. Our framework is stochastic and uses different forms of weak supervision, thus making it possible to train large scale models like deep neural networks.  In summary, we make the following \emph{contributions}:

\begin{itemize}
    \item We define a general framework for adversarial label learning that uses multiple linear constraint forms. 
    \item We develop a stochastic approach that enables training for large-scale models, e.g., deep neural networks. Our algorithm provides greater flexibility by solving a non-zero sum game thus allowing learning of different forms.   
    \item Our evaluation tests weak supervision provided in different forms, either as weak rules from human experts, heuristic patterns, or programmatically generated labels.
\end{itemize}

We validate our approach on three image classification datasets. We use error and precision constraints to solve multiclass image classification tasks for deep neural network models. In the experiments, we provide weak supervision signals that are both generated by humans and programmatically generated. Our results show that our approach outperforms state-of-the-art methods on deep image classification tasks. Our experiments also highlight the difficulties in providing adequate weak supervision signals for solving multiclass image classification tasks.

\section{Related Work}

Our work builds on progress in three topic areas: weak supervision, learning with constraints, and adversarial learning. 

\subsection{Weak Supervision} We expand on some of the recent advances on weakly supervised learning, which is the paradigm where models are trained using large amounts of unlabeled data and low-cost, often noisy annotation. One important recent contribution is the Snorkel system \cite{ratner2017snorkel,ratner2016data,bach2018snorkel}, a weak supervision approach where annotators write different labeling functions that are applied to the unlabeled data to create noisy labels. The noisy labels are combined using a generative model to learn the correlation and dependencies between the noisy signals. Snorkel then reasons with this generative model to produce probabilistic labels for the training data. Our method is related to this approach in that we use noisy labels, or weak signals, to learn adversarial labels for the training data, but our focus is on model training rather than outputting training labels for the unlabeled data. Nevertheless, we show in experiments that the quality of labels learned using Stoch-GALL compares favorably to that of labels inferred by Snorkel's generative modeling. While not the focus of our contributions, the type of human-provided weak signals in our experiments is motivated by techniques in crowdsourcing \cite{gao2011harnessing}.

Weakly supervised methods have enabled knowledge extraction from the Web~\cite{bunescu:acl07,mintz:acl09,riedel:ecml10,yao:emnlp10,hoffman:acl11}, visual image segmentation~\cite{chen:cvpr14,xu:cvpr14}, and tagging of medical conditions from health records~\cite{halpern:health16}. 

\subsection{Learning with Constraints} Our framework incorporates error constraints that are reminiscent of boosting \cite{schapire2002incorporating}; however, our bounds are more general and allow for other forms of constraints like precision. Our work is also related to techniques for estimating accuracies of classifiers using only unlabeled data \cite{blum1998combining,jaffe2016unsupervised,platanios2014estimating,steinhardt2016unsupervised,dawid1979maximum} and combining classifiers for transductive learning using unlabeled data \cite{balsubramani2015optimally,balsubramani2015scalable}. Other methods like posterior regularization (PR) \cite{ganchev2010posterior} and generalized expectation (GE) criteria \cite{druck2008learning,mann2010generalized,mann2008generalized} have been developed to incorporate human knowledge or side information into an objective function. These methods provide parameter estimates as constraints such that the label distributions adhere to these constraints. While GE and PR allow incorporation of weak supervision and parameter estimates as constraints, they do not explicitly consider cases where redundant weak signals that satisfy provided constraints confound the learner. 

\subsection{Adversarial Learning and Games} Researchers have been increasingly interested in adversarial learning \cite{lowd2005adversarial} as a method for training models that are robust to input perturbations of the data. These methods \cite{miyato2018virtual,torkamani2013convex,torkamani2014robustness} regularize the learned model using different techniques to defend against adversarial attacks with an added benefit of improved generalization guarantees. Our approach focuses on adversarial manipulation of the output labels to combat redundancy among multiple sources of weak supervision. 

Game analyses are gaining importance in machine learning because they generalize optimization frameworks by assigning different objective functions for different players or optimizing agents. The generative adversarial network (GAN) \cite{goodfellow2014generative} framework sets up a two player game between a generator and a discriminator, with the aim of learning realistic data distributions for the generator. Our method does not learn a generative model but instead sets up a two-player non-zero sum game between an adversary that assigns labels for the classification task and a model that trains parameters to minimize a cross-entropy loss with respect to the adversarial labels. 

Our work is most closely related to adversarial label learning (ALL) \cite{arachie2019adversarial}, which integrates these topics.
Adversarial label learning was developed for training binary classifiers from weak supervision. The weak supervision was in the form of approximate probabilistic classifications of the unlabeled training data. The algorithm simultaneously optimizes model parameters and estimated labels for the training data subject to the constraint that the error of the weak signals on the estimated labels is within annotator-provided bounds. The ALL paper presents a proof-of-concept that is very limited in its application. ALL requires an error-based loss with error constraints that cannot handle tasks beyond binary classification. 
ALL also uses a learning optimization that requires full gradient updates, restricting its scalability to only small datasets.
We overcome the limitations of ALL by developing a stochastic general framework that enables training of large-scale models and other tasks such as multiclass learning and structured prediction.

\section{Stochastic Generalized Adversarial Label Learning}

Our stochastic general adversarial label learning (Stoch-GALL) approach takes as input an unlabeled dataset and a set of constraints.  
These constraints are consistent with the weak supervision and define the space of possible labelings for the data.
We formulate a nonzero-sum game between two agents: the adversary that optimizes inferred labels for the data and the learner that optimizes parameters for the model. In this game, the objective of the adversary is to assign labels that maximize the error of the model subject to the provided constraints. The model objective is to minimize its loss with respect to the adversarial labels. 
Formally, let the unlabeled training data be $X = \{x_1, \ldots, x_n\}$, and let $f$ be a classifier parameterized by $\theta$.
The primal form of the nonzero-sum game we solve for Stoch-GALL is 
\begin{equation}
\begin{aligned}
\min_{\params} & ~ \loss^{(f_\params)}(X) \text{~~~ and ~~~} \\
\max_{\labelmat \in \Delta^\numdata_\numclasses} & ~  
 \loss^{(\labelmat)}(X) ~~~
\textbf{s.t.} ~~ g_j(\labelmat) \le 0, \forall j \in \{1, \ldots, m\},
\label{eq:primal}
\end{aligned}
\end{equation}
where $\loss^{(f_\params)}$ is a loss function for training the classifier, $\loss^{(\labelmat)}$ is a loss function for the adversarial label perturbations, $\Delta^\numdata_\numclasses$ is the space of \emph{label matrices} where each row is on the simplex of dimension $k$ (i.e., a matrix that can represent a set of $n$ multinomial distributions), $\labelmat$ is the estimated label matrix, $\ell$ is a loss function, and $\{g_1, \ldots, g_m\}$ is a set of linear constraint functions on $\labelmat$. The estimated labels are optimized adversarially ($\max_{Y \in \Delta_C^n}$), against the objective of the learning minimization.

\subsection{Linear Label Constraints}
\label{sec:constraints}

In this section, we describe examples of linear constraints that fit into the Stoch-GALL framework. 
Let $\weakvec \in [0, 1]^n$ be a weak signal that indicates---in a one-versus-rest sense---the probability that each example is in class $c$. And let $\labelvec_c$ denote the $c$th column of matrix $\labelmat$, which is the current label's estimated probability that each example is in class $c$. As originally proposed by \citeauthor{arachie2019adversarial} (\citeyear{arachie2019adversarial}), one set of possible linear constraints that tie the adversarial labels to the true labels is a bound on the error rate of each weak signal.  The expected empirical error for the one-versus-rest task under these two probabilistic label probabilities is
\begin{equation}
\begin{aligned}
\mathrm{error}(\weakvec, \labelvec_c) = \frac{1}{n} \left( \weakvec^\top (1 - \labelvec_c) + (1 - \weakvec)^\top \labelvec_c \right) = \\
\frac{1}{n} \left( \weakvec^\top \boldsymbol{1} + \labelvec_c^\top(1 - 2 \weakvec) \right),
\end{aligned}
\end{equation}
where we use $\weakvec^\top \boldsymbol{1}$ as a vector notation for the sum of $\weakvec$ (or its dot product with the ones vector).
Combined with an annotator provided estimate of a bound ($b_{\text{error}}$) on reasonable errors for their weak signals, an error-based constraint function for weak signal $\weakvec$ on class $c$ would have form
\begin{equation}
    g_{\text{error}}(\labelmat) = \frac{1}{\numdata} \left( \weakvec^\top \boldsymbol{1} + \labelvec_c^\top(1 - 2 \weakvec) \right) - b_{\text{error}}.
\end{equation}

This error-based constraint function can be insufficient to capture the informativeness of a weak signal, especially in cases where there is class imbalance. For multiclass classification, one-versus-rest signals will almost always be class-imbalanced. In such settings, we can allow annotators to indicate their estimates of weak-signal quality by indicating bounds on the precision. In the Stoch-GALL setting, expected precision can also be expressed as a linear function of $\labelmat$:
\begin{equation}
    \mathrm{precision}(\weakvec, \labelvec_c) = \frac{\labelvec_c^\top \weakvec}{\weakvec^\top \boldsymbol{1}}.
\end{equation}
Since $\weakvec$ is a constant with respect to the learning optimization, its appearance in the denominator of this expression does not affect the linearity. We can then define a precision constraint function for each weak signal $\weakvec$ on class $c$:
\begin{equation}
    g_{\text{prec.}}(\labelmat) = b_{\text{prec.}} - \frac{\labelvec_c^\top \weakvec}{\weakvec^\top \boldsymbol{1}}.
\end{equation}
Including precision constraints better captures the confusion matrix across different classes. It is also possible to design other linear constraints.
As long as the constraints are linear, the feasible region for the maximization over $\labelmat$ remains convex. 

\subsection{Nonzero-Sum Losses}

We describe here the loss functions we use to instantiate the Stoch-GALL framework.
The loss functions for the game must be differentiable; but, the choice of loss function is task-dependent and can have important impact on optimization. For multiclass classification using deep neural networks, our model uses popular cross-entropy loss. However, for the adversarial labeling, we instead use an expected error as the loss function, which the adversary maximizes. Formally, the model's loss is the cross-entropy
\begin{equation}
    \loss^{(f_\params)} =  - \frac{1}{n} \sum_{c = 1}^\numclasses \labelvec_c ^\top \log(f_\theta(\Data))
     \label{eq:model_loss},
\end{equation}
while the adversarial labeler's loss is the expected error
\begin{equation}
    \loss^{(\labelmat)} =  \frac{1}{n} \sum_{c = 1}^\numclasses f_\theta(\Data)^\top(1 - \labelvec_c)
     \label{eq:error_loss}.
\end{equation}

We choose this form of error loss because it is linear in the adversarially optimized variable $\labelmat$. This makes the objective for the adversary $\labelmat$ concave, so we are maximizing a concave function subject to linear constraints. This makes the adversarial optimization a linear program with a unique optimum for any fixed $f_\params$. This form is relevant for the initialization scheme described in \cref{sec:optimization}. We optimize the loss functions using Adagrad \cite{duchi2011adaptive}.


\subsection{Optimization}
\label{sec:optimization}

We use a primal-dual optimization that jointly solves an augmented Lagrangian relaxation of \cref{eq:primal}. Since our formulation uses a nonzero-sum game, we have two separate optimizations. The analogous optimizations are
\begin{equation}
\begin{aligned}
    \min_{\params} & ~ \loss^{(f_\params)} \text{~~~ and ~~~} \\
    \min_{\lagrangevec \in \reals^\numconstraints_+} & ~ \max_{\labelmat \in \Delta^\numdata_\numclasses} ~  
     \loss^{(\labelmat)} - \lagrangevec^\top G(\labelmat) - \frac{\stepsize}{2}\left|\left| G(\labelmat) \right|\right|^2_{+} := \hat{\loss}^{(\labelmat)}~,
     \label{eq:lagrangian}
\end{aligned}
\end{equation}
where $\lagrangevec$ is the vector of Karush-Kuhn-Tucker (KKT) multipliers, $G$ is the vector of constraint function outputs (i.e., $G(\labelmat)_j := g_j(\labelmat)$), $\stepsize$ is a positive parameter, and $|| \cdot ||^2_+$ denotes the norm of positive terms. The adversary optimization maximizes the linear loss from \cref{eq:error_loss} while the learner minimizes the model loss from \cref{eq:model_loss}.  A primal-dual solver for this problem updates the free variables using interleaved variations of gradient ascent and descent. We do this by updating $\labelmat$ and $\lagrangevec$ with their full gradients while holding $\theta$ fixed. Afterwards, we fix $\labelmat$ and $\lagrangevec$ and update $\theta$ with its mini-batches for a fixed number of epochs. The number of epochs depends on the size and architecture of the model network.

To preserve domain constraints on the variables $\labelmat \in \Delta^\numdata_\numclasses$ and $\lagrangevec \ge 0$, we use projection steps that enforce feasibility. After each update to $\labelmat$ and $\lagrangevec$, we project $\labelmat$ to the simplex using the sorting method \cite{blondel2014large}, and we clip $\lagrangevec$ to be non-negative.

\subsubsection{Initialization Scheme} To further facilitate faster convergence toward a local equilibrium, we warm start the optimization with a phase of optimization updating only $\labelmat$ and $\lagrangevec$. The effect of this warm-start phase is that we begin learning with a near-feasible $\labelmat$---one that is nearly consistent with the weak-supervision-based constraints. Since this phase uses the fixed output of a randomly initialized model $f_\params$, it does not require forward- or back-propagation through the deep neural network, so it is inexpensive, even for large datasets.

\subsubsection{Analysis}

The advantage of this primal-dual approach is that it enables inexpensive updates for the gaming agents and other variables being optimized, thereby allowing learning to occur without waiting for the solution of the inner optimization. At every iteration, the primal variables take maximization steps and the dual variables take minimization steps. However, for training deep neural networks, the primal-dual approach is not always ideal.

The large datasets needed to fit large models such as deep neural networks often require stochastic optimization to train efficiently. The key computational benefit of stochastic optimization is that it avoids the $O(n)$ cost of computing the true gradient update. Using a primal-dual approach to optimize \cref{eq:lagrangian} would also incur an $O(n)$ cost for each update to $\labelmat$. This cost is why we design our optimization scheme to update $\labelvec$ and $\lagrangevec$ only after a fixed number of epochs. Since each epoch costs $O(n)$ computation, the added overhead does not change the asymptotic cost of training.

This optimization scheme has an added benefit that it increases the stability of the learning algorithm. By updating $\labelmat$ and $\lagrangevec$ only after a few epochs of training $\params$, we are solving the minimization over $\params$ nearly to convergence. We still retain the advantages of primal-dual optimization over the $\labelmat$ variables, but without the added instability of simultaneous nonconvex optimization.

\begin{algorithm*}[htb]
\begin{algorithmic}[1]
\REQUIRE Dataset $\Data = [\data_1, \ldots, \data_\numdata]$, vector of constraint functions $G$, augmented Lagrangian parameter $\stepsize$.
\STATE Initialize model parameters $\params$ (e.g., deep neural network weights)
\STATE Initialize $\labelmat \in \Delta^\numdata_\numclasses$ (e.g., uniform probability)
\STATE Initialize $\lagrangevec \in \reals_{\ge 0}$ (e.g., zeros)
\WHILE{$G(\labelmat) > $ tolerance}
\STATE Update $\labelmat$ with gradient $\nabla_{\labelmat} \hat{\loss}^{(\labelmat)}$ (e.g., $\labelmat \leftarrow \labelmat + \alpha \nabla_{\labelmat} \hat{\loss}^{(\labelmat)}$)
\STATE Project $\labelmat$ to $\Delta^\numdata_\numclasses$
\STATE Update $\lagrangevec$ with gradient $\nabla_{\lagrangevec} \hat{\loss}^{(\labelmat)}$ (e.g., $\lagrangevec \leftarrow \lagrangevec - \stepsize G(\labelmat)$) 
\STATE Clip $\lagrangevec$ to be non-negative
\ENDWHILE
\WHILE{$\params$ not converged}
\STATE Update $\params$ with $\nabla_{\params} L^{(f_\params)}_\mathcal{B}$ (mini-batches) for a fixed number of epochs
\STATE Update $\labelmat$ with gradient $\nabla_{\labelmat} \hat{\loss}^{(\labelmat)}$
\STATE Project $\labelmat$ to $\Delta^\numdata_\numclasses$
\STATE Update $\lagrangevec$ with gradient $\nabla_{\lagrangevec} \hat{\loss}^{(\labelmat)}$
\STATE Clip $\lagrangevec$ to be non-negative
\ENDWHILE
\RETURN model parameters $\params$
\end{algorithmic}
\caption{Stochastic Generalized Adversarial Label Learning}
\label{alg:ALL}
\end{algorithm*}

\section{Experiments}
\label{sec:experiments}

We validate our approach on three fine-grained image classification tasks, comparing the performance of models trained with our approach to a baseline averaging method and model trained using labels generated from Snorkel \cite{ratner2017snorkel}. Each of these methods trains from weak signals, and our experiments evaluate how well they can integrate noisy signals and how robust they are to confounding signals.

\subsection{Quality of Constrained Labels}

Before using our custom weak annotation framework, we first compare the quality of labels generated by Stoch-GALL to existing methods for fusing weak signals. We follow the experiment design from a tutorial\footnote{\url{https://github.com/HazyResearch/snorkel/blob/master/tutorials/images/Images_Tutorial.ipynb}} designed by \citeauthor{ratner2017snorkel} (\citeyear{ratner2017snorkel}) to demonstrate their Snorkel system's ability to fuse weak signals and generate significantly higher quality labels than naive approaches. 
The experiment uses \emph{Microsoft COCO: Common Objects in Context} \cite{plummer2015flickr30k} dataset to train detectors of whether a person is riding a bike within each image. We use the 903 images from the tutorial and weak signals generated by the labeling functions based on object occurrence metadata. 
We calculate the error and precision of each rule and use those to define Stoch-GALL constraints. We then run the initialization scheme (the first while loop in \Cref{alg:ALL}), which finds feasible labels adversarially fit against a random initialization, i.e., arbitrary feasible labels. For an increasing number of weak signals, we compare ALL with error constraints, Stoch-GALL with both error and precision constraints, Snorkel, and majority voting.

\begin{figure}[htb]
\centering
\includegraphics[width=0.5\textwidth]{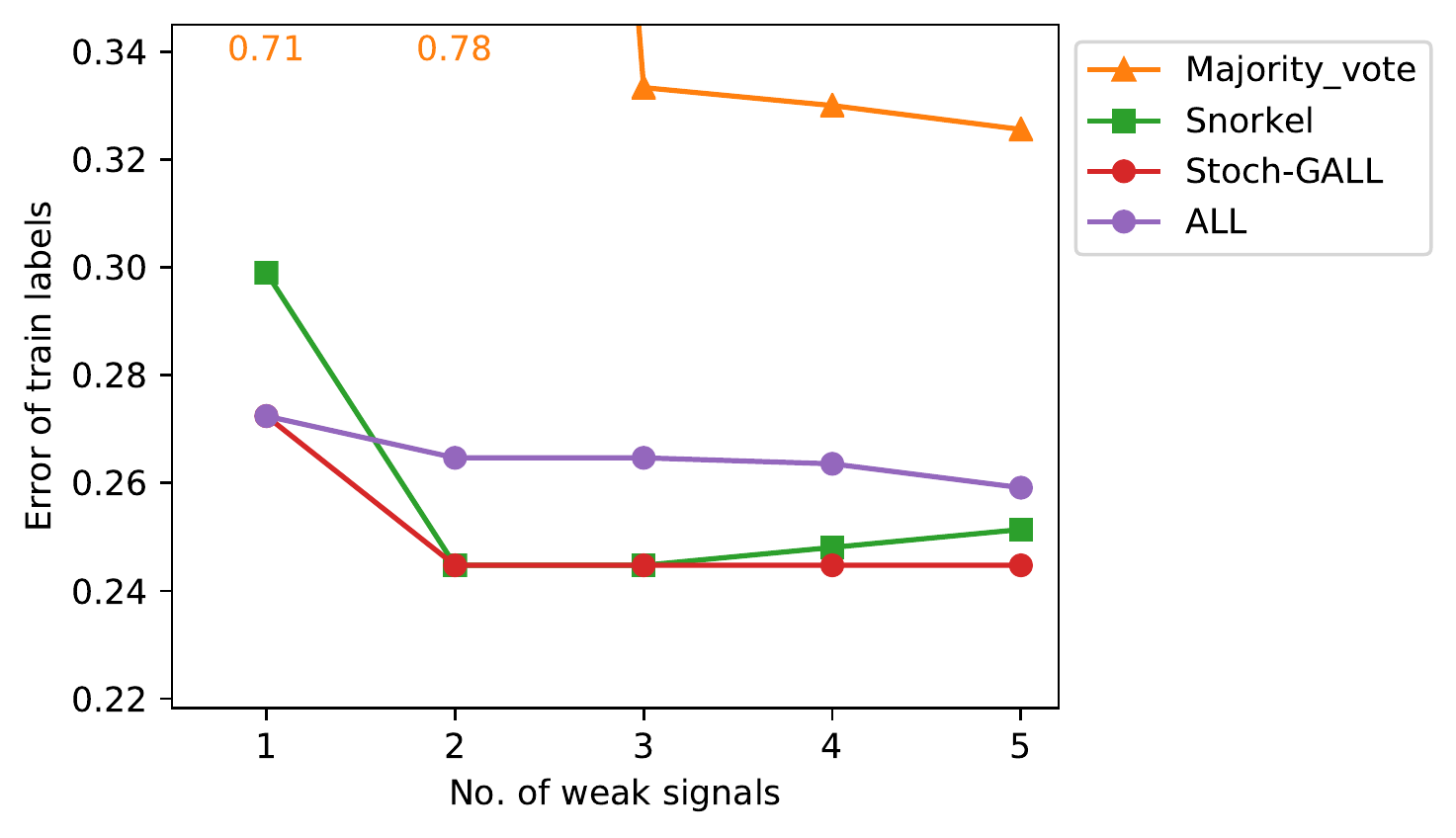}
\caption{Error of MSCOCO bike-riding labels using Stoch-GALL initialization compared to other methods.}
\label{fig:label-quality}
\end{figure}

We plot the resulting error rate of the generated labels in \Cref{fig:label-quality}. For all numbers of weak signals, Stoch-GALL obtains the highest accuracy labels. The labels generated by Snorkel have the same label error using two and three weak signals, but adding additional weak signals starts to confound Snorkel. Our framework is not confounded by these additional weak signals. Finally, corroborating the results reported by Snorkel's designers, the naive majority vote method has significantly higher error compared to any of the more sophisticated weak supervision techniques.

\subsection{Multiclass Image Classification}

For our other experiments, we train multiclass image classifiers from weak supervision. We are interested in evaluating the effectiveness of the weak supervision approach, so we use the same deep neural network architecture for all experiments: a six-layer convolutional neural network where each layer contains a max‐pooling unit, a relu activation unit, and dropout. The final layer is a fully connected layer with softmax output. 
\Cref{tab:scores} lists summary results of the error obtained by each method on each dataset using all the weak signals we provide the learners. The final result for each experiment is that Stoch-GALL outperforms both Snorkel and averaging in all settings, showing a strong ability to fuse noisy signals and to avoid being confounded by redundant signals.
We describe our form of weak supervision and each experiment in detail in the rest of the section.

\subsubsection{Weak Signals}
\label{weaksignal}

We ask human annotators to provide weak signals for image datasets. To generate each weak signal, we sample 50 random images belonging to different classes. We then ask the annotators to select a representative image and mark distinguishing regions of the image that indicate its belonging to a specific class. We then calculate pairwise comparisons between the pixels in the region of the reference image selected by an annotator and the pixels in the same region for all other images in the dataset. We measure the Euclidean distance between the pairs of images and convert the scores to probabilities with a logistic transform. Through this process, an annotator is guiding the design of simple nearest-neighbor one-versus-rest classifiers, where images most similar to the reference image are more likely to belong to its class. We ask annotators to generate many of these rules for the different classes, and we provide the computed probabilities as weak labels for the weakly supervised learners. 

In practice, we found that these weak signals were noisy. In some experiments, they were insufficient to provide enough information for the classification task. However, our experiments show how different weakly supervised learners behave with informative but noisy signals. We discuss ideas on how to design better interfaces and better weak signals for the image classification task in \cref{sec:discussion}.

We assume we have access to a labeled validation set consisting of 1\% of the available data. We use this validation set to compute the precision and error bounds for the weak signals. 
This validation set is meant to simulate a human expert's estimate of error and precision.
To encourage a fair comparison, we allow all methods to use these labels in addition to weak signals when training by appending the validation set to the dataset with its true labels. 
Since these bounds are evaluated on a very tiny set of the training data, they are noisy and prone to the same type of estimation mistakes an expert annotator may make. Therefore, they make a good test for how robust Stoch-GALL is to imperfect bounds. 

\begin{table*}[tb]
    \centering
    \begin{tabular}{l r r r}
        \toprule
        Data & Stoch-GALL & Average & Snorkel\\
        \midrule
        Fashion-mnist (weak) & \textbf{0.335} & 0.447 & 0.401  \\
        Fashion-mnist (pseudolabels + weak) & \textbf{0.228} & 0.315 & 0.320 \\
        SVHN (pseudolabels + weak) & \textbf{0.231} & 0.435 & 0.525 \\
        \bottomrule
    \end{tabular}
    \caption{Errors of models trained using all weak signals. In all three settings, Stoch-GALL is able to train higher accuracy models than Snorkel or average labels. The settings include using all human-annotated weak signals (weak) and combining the human signals with pseudolabels (pseudolabels + weak).}
    \label{tab:scores}
\end{table*}

\subsubsection{Weakly Supervised Image Classification}

In this experiment, we train a deep neural network using only human-provided weak labels as described in \cref{weaksignal}. We use the \emph{Fashion-mnist} \cite{xiao2017fashion} dataset, which represents an image-classification task where each example is a $28 \times 28$ grayscale image. The images are categorized into 10 classes of clothing types with 60,000 training examples and 10,000 test examples. We have annotators generate five one-versus-rest weak signals for each class, resulting in 50 total weak signals.

\begin{figure}[tb]
\centering
\begin{subfigure}{.45\textwidth}
\includegraphics[width= \textwidth]{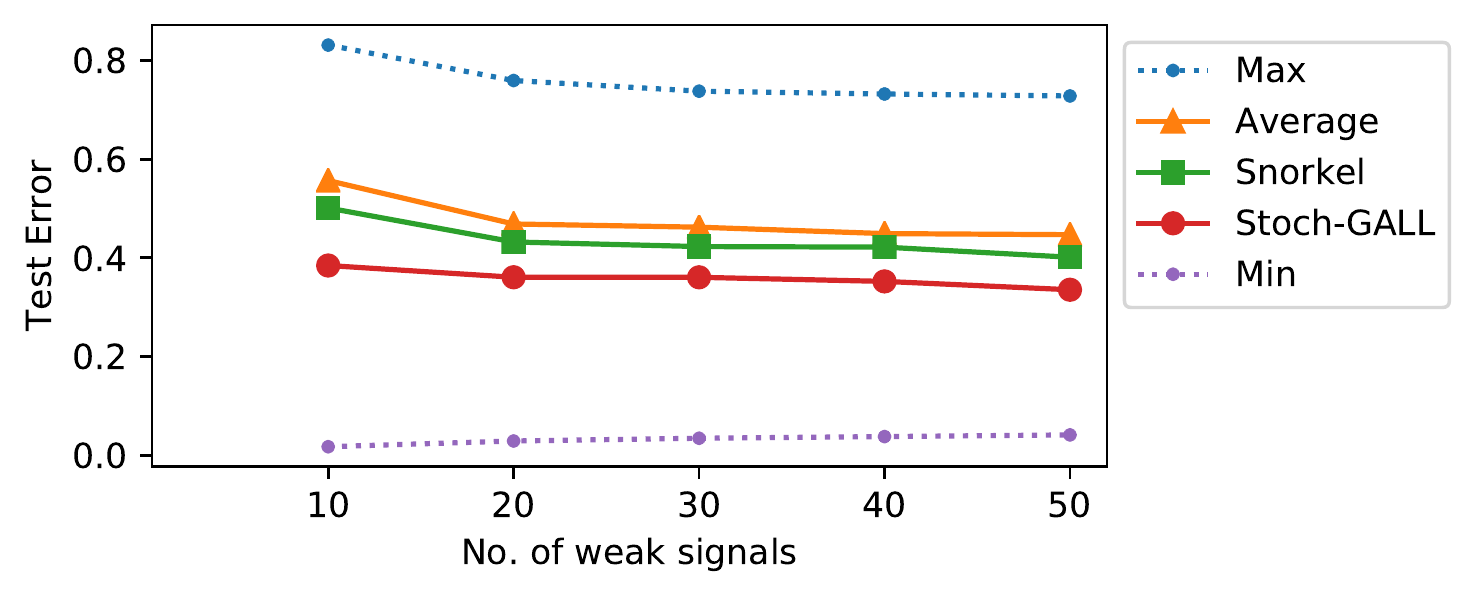}
\caption{Test error}
\label{fig:fmnist-weak}
\hfill
\end{subfigure}
\begin{subfigure}{.35\textwidth}
\includegraphics[width= \textwidth]{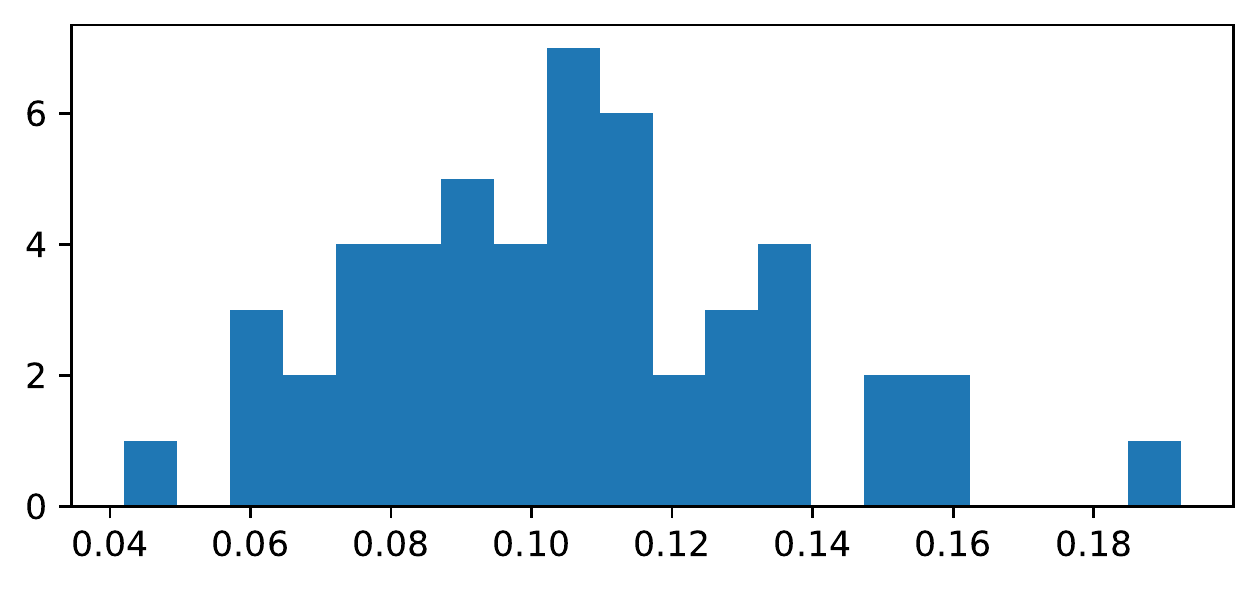}
\caption{Error bounds}
\label{fig:fmnist_weakerror}
\end{subfigure}
\hfill
\begin{subfigure}{.35\textwidth}
\includegraphics[width= \textwidth]{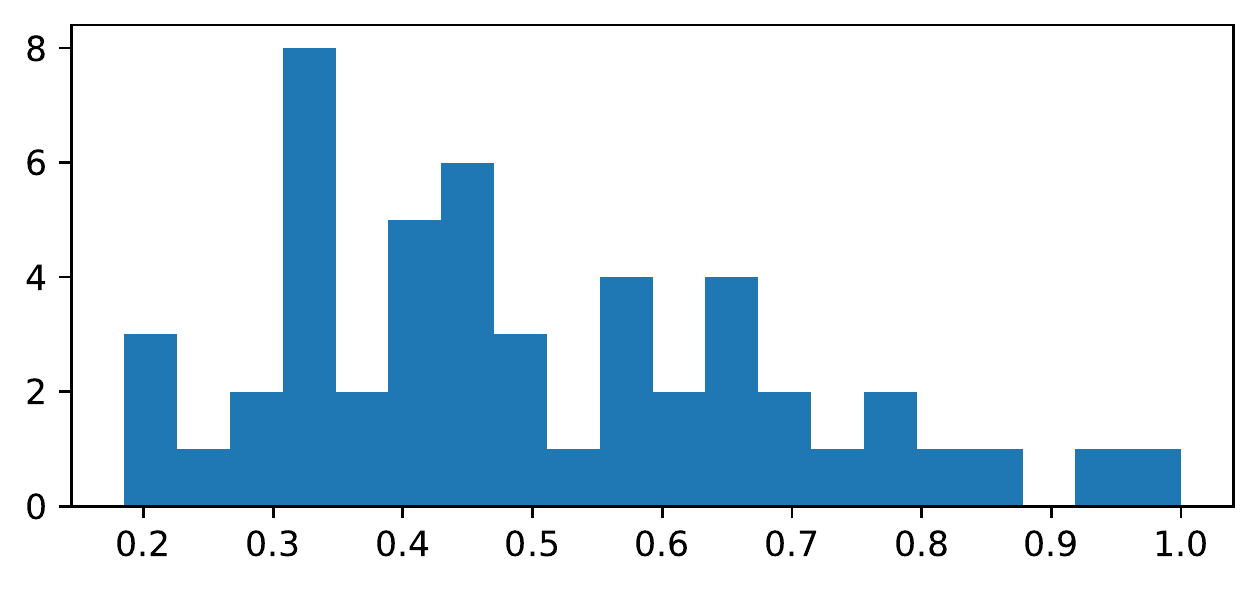}
\caption{Precision bounds}
\label{fig:fmnist_weakprec}
\end{subfigure}
\caption{Analyses of experiments using the Fashion-mnist dataset with human-provided weak signals.}
\label{fig:fashionmnistw}
\end{figure}

We plot analyses of models trained using weak supervision in \Cref{fig:fashionmnistw}, where \cref{fig:fmnist-weak} plots the test error, and \cref{fig:fmnist_weakerror} and \cref{fig:fmnist_weakprec} are histograms of the error and precision bounds for the weak signals evaluated on the validation set. Since our weak signal is a one-versus-rest prediction of an image belonging to a particular class, the baseline precision and error should be 0.1 for training data with balanced classes. The histograms indicate that there is a wide range of precisions and errors for the different weak signals. Note that the order of the weak signals in our experiment was fixed as the order provided by the annotators. Thus, the first weak signal for each task is the first weak signal that the annotator generated for that dataset.
The error rates in \cref{fig:fmnist-weak} suggest that the test error of the models decreases as we add more weak signals.  Stoch-GALL with both precision and error bounds outperforms Snorkel and the average baseline for all the weak signals. The min and max curves in the plots represent the best and worst possible label errors for labels that satisfy the provided constraints. The high error in the max curve indicates that the constraints alone still allow highly erroneous labels, yet our Stoch-GALL framework trains models that perform well. The min curve indicates how close feasible adversarial labels could be to the true labels. In this experiment, the min curve is close to zero, which suggests that the inaccuracies in the provided bounds are not overly restrictive. 

\subsubsection{Weak-Pseudolabel Image Classification}

In the previous experiments, the human provided weak signals are informative enough to train models to perform better than random guessing, but the resulting error rate is still significantly lower than that of supervised methods. To further boost the performance, we combine the human weak signals with \emph{pseudolabels}: predictions of our deep model trained on the validation set and applied to the unlabeled training data \cite{lee2013pseudo}. By training on the available 1\% labels and predicting labels for the remaining 99\% unlabeled examples, we create a new, high-quality weak signal. We calculate error and precision bounds for the pseudolabels with four-fold cross-validation on the validation set. We report the results of the models trained on the Fashion-mnist dataset using this combination of pseudolabels and human weak signals.

\begin{figure}[tb]
\centering
\begin{subfigure}{.45\textwidth}
\includegraphics[width= \textwidth]{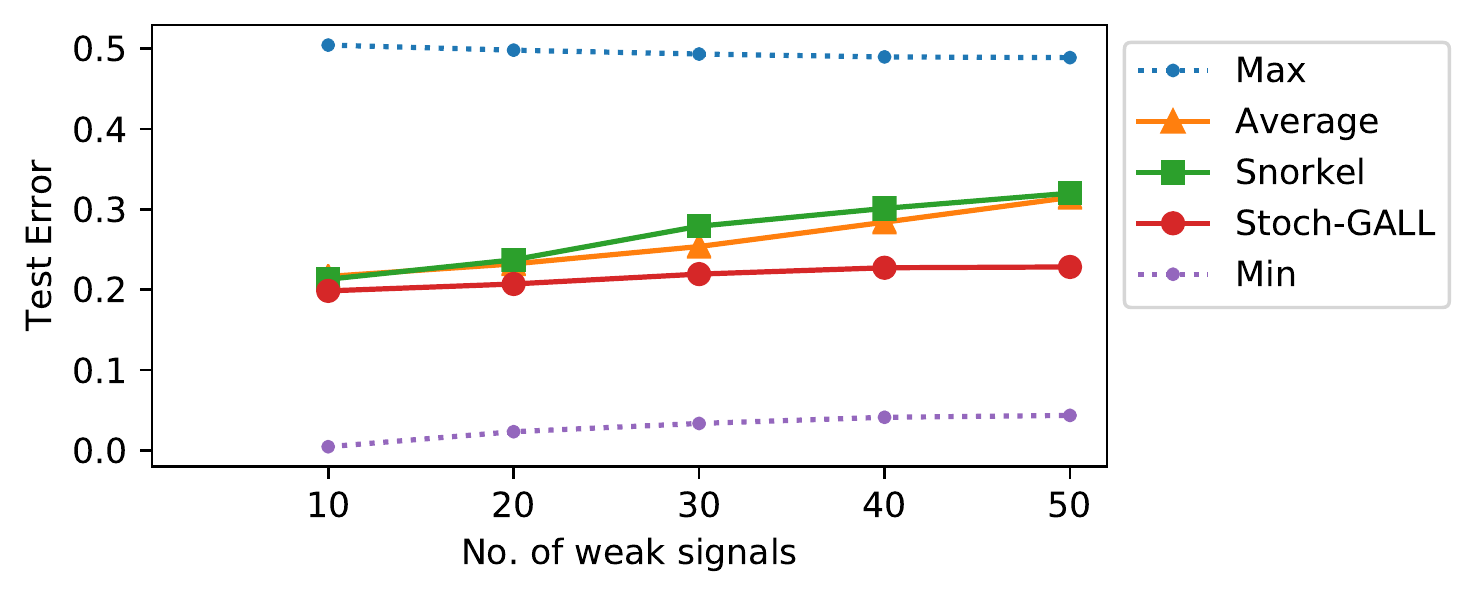}
\caption{Test error}
\label{fig:fmnist-semi}
\hfill
\end{subfigure}
\begin{subfigure}{.35\textwidth}
\includegraphics[width= \textwidth]{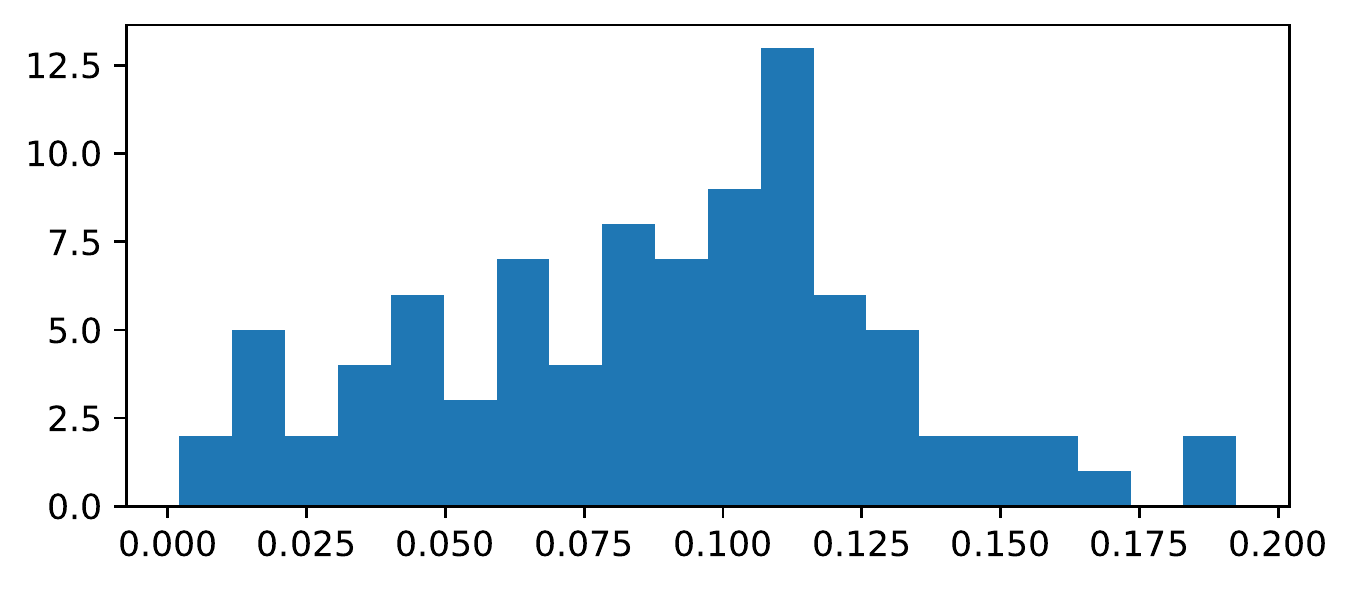}
\caption{Error bounds}
\label{fig:fmnist_semierror}
\end{subfigure}
\hfill
\begin{subfigure}{.35\textwidth}
\includegraphics[width= \textwidth]{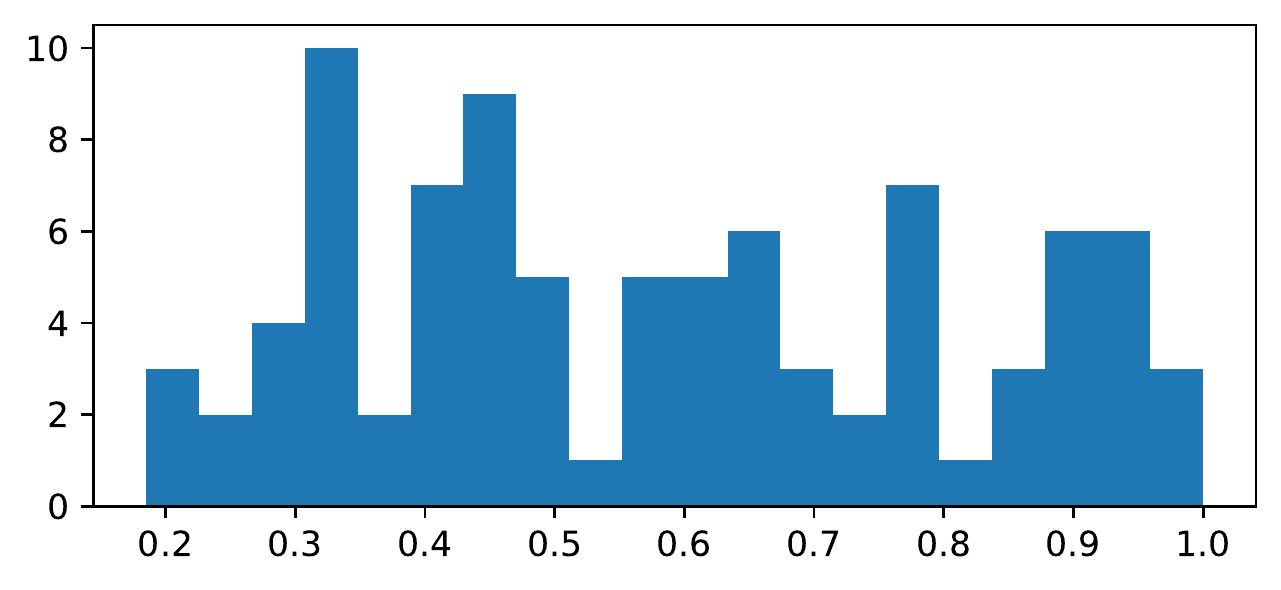}
\caption{Precision bounds}
\label{fig:fmnist_semiprec}
\end{subfigure}
\caption{Test error on Fashion-mnist dataset using pseudolabels and human weak signals.}
\label{fig:fashionmnists}
\end{figure}

\cref{fig:fashionmnists} contains plots of the results. The error and precision histograms now include higher precision bounds and lower error bounds, as a result of the pseudolabel signals being higher quality than the human provided weak signals. Additionally, the min and max error curves have lower values, indicating that we get better quality labels with these signals. The error trends in \cref{fig:fmnist-semi} are quite different compared to the previous experiment (\cref{fig:fmnist-weak}). All the methods have good performance with the pseudolabel signals, but as we add the human signals, Snorkel and the average baseline are confounded and produce increasingly worse predictions. Stoch-GALL however is barely affected by the human weak signals. The slight variation in the curve can be attributed to the inaccuracy of estimated bounds for the weak signals.

We hypothesize that this trend occurs because of the nature of our weak supervision. Since the weak signals are based on the selection of exemplar images, they may be effectively subsumed by a fully semi-supervised approach such as pseudolabeling. That is, the information provided by each human weak signal is already included in the pseudolabeling signal. This type of redundant information is common when using weak supervision. Many signals can have dependencies and redundancies. And despite the Snorkel system's modeling of dependencies among weak signals, it is still confounded by them while Stoch-GALL's model-free approach is robust.

Our final experiments test the performance of the different models using pseudolabels and human weak labels on another image classification task. We use the Street View House Numbers (SVHN) \cite{netzer2018street} dataset, which represents the task of recognizing digits on real images of house numbers taken by Google Street View. Each image is a $32 \times 32$ RGB vector. The dataset has 10 classes consisting of 73,257 training images and 26,032 test images.

\begin{figure}[tb]
\centering
\begin{subfigure}{.45\textwidth}
\includegraphics[width= \textwidth]{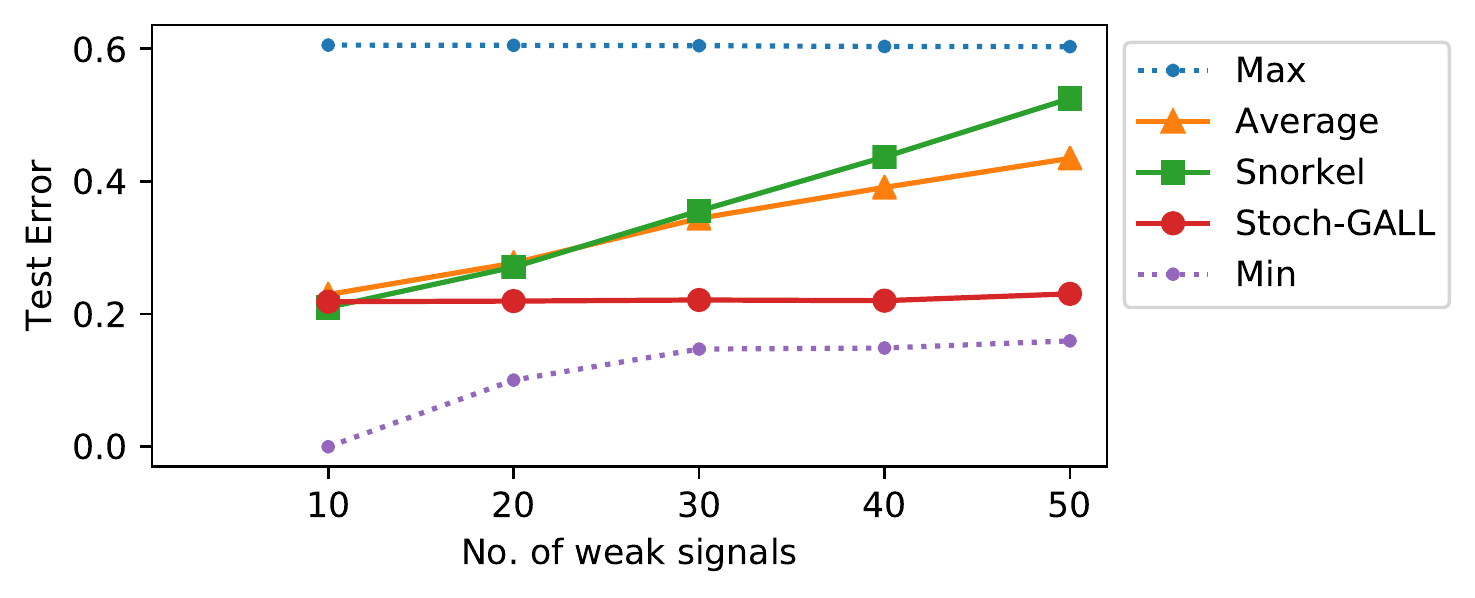}
\caption{Test error}
\label{fig:svhn-semi}
\hfill
\end{subfigure}
\begin{subfigure}{.35\textwidth}
\includegraphics[width= \textwidth]{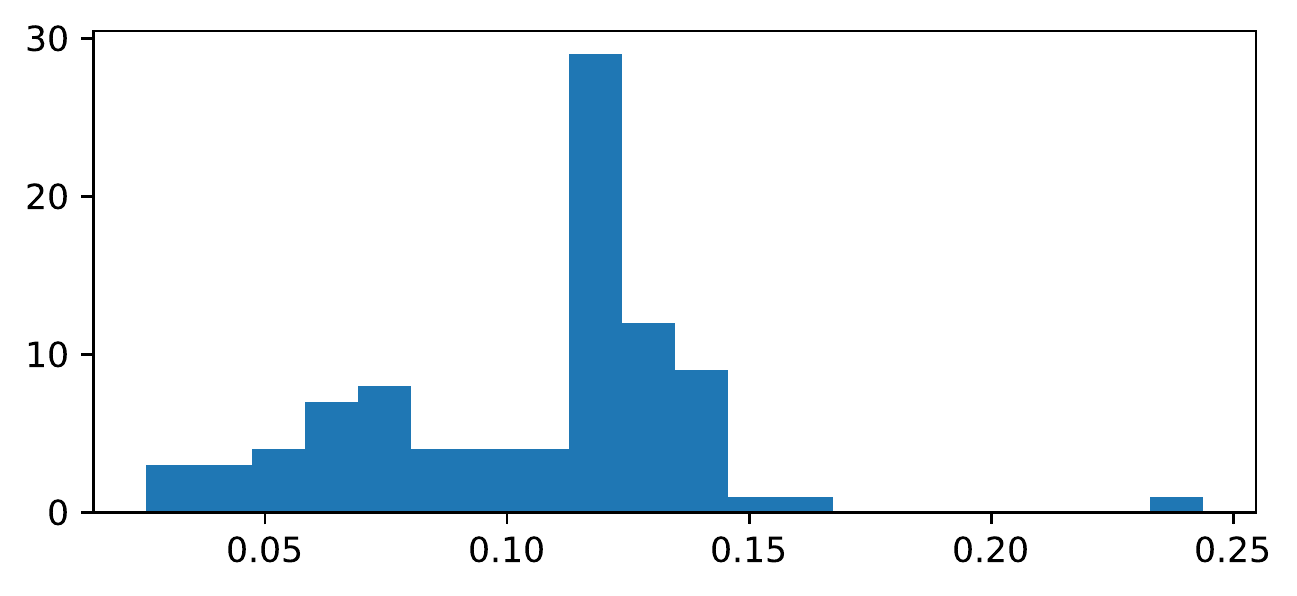}
\caption{Error bounds}
\label{fig:sevhn_error}
\end{subfigure}
\hfill
\begin{subfigure}{.35\textwidth}
\includegraphics[width= \textwidth]{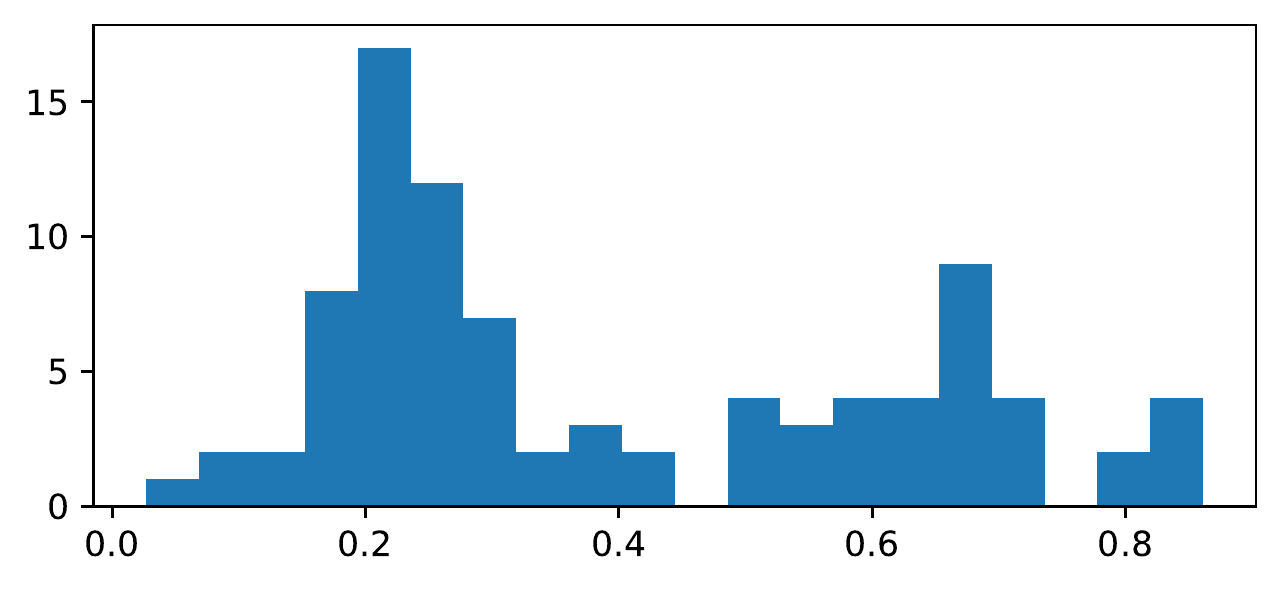}
\caption{Precision bounds}
\label{fig:svhn_prec}
\end{subfigure}
\caption{Test error on SVHN dataset using pseudolabels and human weak signals.}
\label{fig:svhn}
\end{figure}

\Cref{fig:svhn} plots the results of the experiment. \Cref{fig:svhn-semi} features the same trend as \cref{fig:fmnist-semi}. For this task, the human weak signals perform poorly in labeling the images, so they do not provide additional information to the learners. This fact is evident in the horizontal slope of the max curve. The min curve suggests that the human weak signals are redundant with poorly estimated bounds, and adding them decreases the space of possible labels for Stoch-GALL. Comparing models, Stoch-GALL's performance is not affected by the redundancies in the weak signals. Since the human weak signals are very similar, Snorkel seems to mistakenly trust the information from these signals more as we add more of them, thus hurting its model performance. Stoch-GALL uses the extra information provided to it as bounds on the weak signals to protect against placing higher emphasis on the redundant human weak signals.

\section{Discussion}
\label{sec:discussion}

We introduced a stochastic generalized adversarial label learning framework that enables users to encode information about the data as a set of linear constraints. We show in our experiments the performance of the method using precision and error constraints. However, Stoch-GALL allows for other forms of linear constraints. Our evaluation demonstrates that our adaptive framework is able to generate high-quality labels for a learning task and is also able to combine different sources of weak supervision to increase the performance of a model. Our experiments show that our framework outperforms state-of-the-art weak supervision method on different image-classification tasks and is better at handling redundancies among weak supervision signals. 

In our work, we used simple nearest-neighbor weak signals provided by human annotators and programmatically generated weak signals. The human provided weak signals performed well in some experiments, but in other experiments, they did not provide adequate information to the learner. In future work, we will explore avenues for generating higher quality human supervision signals. One idea for improving the human signals is by learning latent data representations (e.g., with an autoencoder) and comparing the latent representations, rather than raw pixel values. Since our annotators have no expertise about the data, we simulated expert knowledge by evaluating the weak signals on a tiny set of the training data. We can also estimate these bounds using agreements and disagreements of the weak signals, but this will involve advanced modeling techniques. Lastly, although we have only shown results for multi-classification on image datasets, users can use our framework for other forms of learning such as multilabel classification or structured prediction by using different loss functions.

\section*{Acknowledgments}

We thank NVIDIA for their support through the GPU Grant Program and Amazon for their support via the AWS Cloud Credits for Research program.

\bibliographystyle{aaai}
\bibliography{arachie}

\end{document}